\documentclass[journal]{IEEEtran}

\usepackage{cite}

\usepackage{amsmath}
\usepackage{algorithm}
\usepackage{algpseudocode}
\usepackage{makecell}

\usepackage{float}
\usepackage{subcaption}
\usepackage{booktabs} % For improved table formatting
\usepackage{array}    % For better array/table formatting
\usepackage{caption}  % For caption control
\usepackage{multirow} % Optional, in case you need to merge rows
\usepackage{color}

\ifCLASSINFOpdf
  \usepackage[pdftex]{graphicx}
  \graphicspath{{../pdf/}{../jpeg/}}
  \DeclareGraphicsExtensions{.pdf,.jpeg,.png}
\else

\fi

\usepackage{makecell}
\usepackage{multirow} % For multirow/column cells
\usepackage{booktabs} % For better table formatting
\usepackage{array}

\usepackage{array}

\begin{document}

\title{Visual Environment-Interactive Planning for Embodied Complex-Question Answering}

\author{Ning Lan, Baoshan Ou, Xuemei Xie,~\IEEEmembership{Senior Member,~IEEE} and Guangming Shi,~\IEEEmembership{Fellow,~IEEE}% <-this % stops a space

\thanks{Received 11 October 2024; revised 28 December 2024; accepted 2 February }
\thanks{Ning Lan and Baoshan Ou are with the School of Artificial Intelligence, Xidian University, Xi’an 710071, China (e-mail: lann@stu.xidian.edu.cn; bsou@stu.xidian.edu.cn).}% <-this % stops a space
\thanks{Xuemei Xie is with the School of Artificial Intelligence, Xidian University, Xi’an 710071, China, also with Guangzhou Institute of Technology, Xidian University, Guangzhou 510000, China, and also with Pazhou Laboratory, Huangpu, Guangzhou 510000, China (e-mail: xmxie@mail.xidian.edu.cn).}% <-this % stops a space
\thanks{Guangming Shi is with the School of Artificial Intelligence, Xidian University, Xi’an 710071, China, and also with the Peng Cheng Laboratory, Shenzhen 518055, China (e-mail: gmshi@mail.xidian.edu.cn).}% <-this % stops a space
\thanks{Digital Object Identifier 10.1109/TCSVT.2025.3538860}}% <-this % stops a space

\markboth{IEEE TRANSACTIONS ON CIRCUITS AND SYSTEMS FOR VIDEO TECHNOLOGY}%
{Shell \MakeLowercase{\textit{et al.}}: Bare Demo of IEEEtran.cls for IEEE Journals}

\IEEEpubid{\begin{minipage}{\textwidth}\ \\[12pt] \centering
	1051-8215 © 2025 IEEE. All rights reserved, including rights for text and data mining, and training of artificial intelligence \\
	and similar technologies. Personal use is permitted, but republication/redistribution requires IEEE permission.\\
	See https://www.ieee.org/publications/rights/index.html for more information.
\end{minipage}}

\maketitle

\begin{abstract}
This study focuses on Embodied Complex-Question Answering task, which means the embodied robot need to understand human questions with intricate structures and abstract semantics. The core of this task lies in making appropriate plans based on the perception of the visual environment. Existing methods often generate plans in a \textit{once-for-all} manner, $i.e.$, \textit{one-step planning}. Such approach rely on large models, without sufficient understanding of the environment. Considering \textit{multi-step planning}, the framework for formulating plans in a \textit{sequential} manner is proposed in this paper. To ensure the ability of our framework to tackle complex questions, we create a structured semantic space, where hierarchical visual perception and chain expression of the question essence can achieve iterative interaction. 
This space makes \textit{sequential} task planning possible. Within the framework, we first parse human natural language based on a visual hierarchical scene graph, which can clarify the intention of the question. Then, we incorporate external rules to make a plan for current step, weakening the reliance on large models. Every plan is generated based on feedback from visual perception, with multiple rounds of interaction until an answer is obtained. This approach enables continuous feedback and adjustment, allowing the robot to optimize its action strategy. To test our framework, we contribute a new dataset with more complex questions. Experimental results demonstrate that our approach performs excellently and stably on complex tasks. And also, the feasibility of our approach in real-world scenarios has been established, indicating its practical applicability.

\end{abstract}


\IEEEpeerreviewmaketitle

\section{Introduction}

\IEEEPARstart {T}{he} development of versatile embodied agents capable of understanding natural language commands in indoor environments and executing various tasks through visual interaction has been a long-standing goal. This research area, known as Embodied Question Answering (EmbodiedQA)\cite{das2018embodied} tasks, involves robots equipped with visual and auditory sensors that navigate indoor spaces based on human instructions \cite{zhan2024enhancing}\cite{wang2023res} and provide answers through visual observation\cite{chen2023think}. For example, when given the instruction, ``Is there a phone on the table in the living room?'', the robot must comprehend the intent of the question to formulate plans, autonomously explore the environment to locate the target object, and then integrate information from both visual and linguistic sources to generate an accurate response. As the complexity and scope of task instructions increase, the difficulty of task planning also rises. Here we extend EmbodiedQA to embodied complex-question answering tasks, which are closer to real human-robot interaction and better reflect the robot's ability to plan complex tasks. 

Recently, large language models (LLMs) have found increasing applications in the field of embodied robots\cite{guo2024embodied}\cite{majumdar2024openeqa}\cite{rana2023sayplan}\cite{birr2024autogpt+}\cite{liu2024llm}. These studies effectively utilize the abstract reasoning capabilities of LLMs to assist robots in task planning. However, the application of LLMs-based embodied robots for embodied complex-question answering faces significant challenges and barriers. Existing LLMs-based task planning methods typically generate plans in a one-shot manner. These methods often lack sufficient understanding of the environment, making them inadequate for handling complex situations. They are highly susceptible to the hallucinations of the LLMs, leading to unreliable task execution outcomes. Meanwhile, due to the lack of interpretability in LLMs, it is challenging to pinpoint where errors occur, making problem diagnosis and correction more difficult. 

\begin{figure*}[htbp]
	\centerline{
				\includegraphics[width=18cm]{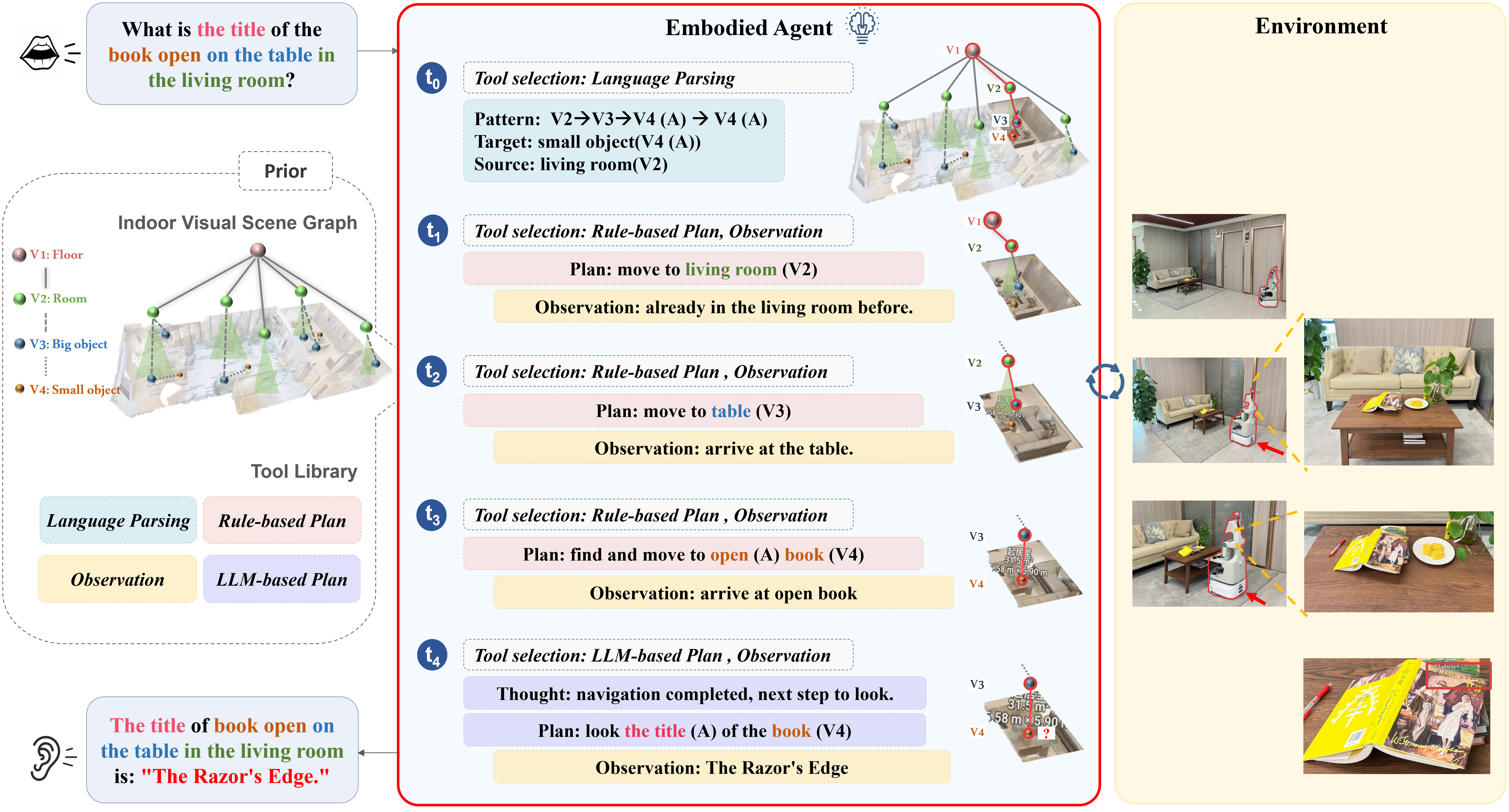} 	
					}
	\caption{An illustration of our framework for task planning. The embodied agent is the core the system. Inside the ``Embodied Agent'' block, we construct a structured semantic space, enabling continuous interaction between natural language instructions and visual perception. After receiving the natural language question, the embodied agent first select the Language Parsing tool to obtain the pattern of question($t_0$). The intent is to obtain the attribute ($A$) of a book in the small layer ($V_4$). By combining the pattern with rule-based analysis, the optimal observation point to answer the question is at the book of small object layer. Therefore, the embodied agent next select Rule-based Plan and Observation tools to formulate current plan for navigation by interacting with the environment($t_1$, $t_2$, $t_3$). When the embodied agent determines that it cannot give a plan based on rules, it will choose the LLM-based Plan tool, utilizing the powerful language ability of LLMs to give plans based on feedback from environmental interactions($t_4$).}
	\label{framework}
\end{figure*}

In this paper, we propose an embodied agent framework that formulates plans in a sequential manner. As illustrated in Fig. \ref{framework}, the human instruction ($i.e.$ language modality) is on the left, and the observation feedback from the environment ($i.e.$ visual modality) is on the right. The embodied agent is the core of the system. Inside the ``Embodied Agent'' block, the right side highlights the interaction process, where different modalities interact within the structured semantic space. The visual modality is represented as a hierarchical graph structure in this space, including four layers: floor, room, big object, and small object. The linguistic modality is expressed as a chain structure, which is attached to the visual scene graph through red lines connection. The whole interaction processes follow a top-down manner, refining information step by step from a coarse-grained to a fine-grained level.
	
Specifically, the indoor visual scene graph and the tool library are pre-constructed and serve as prior knowledge to assist the robot in each step of the planning process. Upon receiving intricate human instructions ($e.g.$, ``What is the title of the book open on the table in the living room?''), 
\IEEEpubidadjcol
the robot first employs the Language Parsing tool to project the intent of instructions into the structured semantic space. Then combined with the visual semantics perceived in the previous step, it utilizes the Rule-based planning tool or the LLM-based planning tool to formulate a plan for the current step. After executing the plan, the robot uses the Observation tool to perceive the environment and confirm whether the task has been successfully completed. By sequentially developing plans through the interaction of visual environment and language expression, robots can complete complex tasks through gradual exploration.

To better understand questions and formulate plans incrementally, the embodied agent needs to have a fundamental prior understanding of the visual scene. This paper first constructs the indoor visual scene graph, which can be seen as a carrier of structured semantic space. The graph is hierarchically structured as ``floor layer($V_1$) $\rightarrow$ room layer($V_2$) $\rightarrow$ big object layer($V_3$) $\rightarrow$ small object layer($V_4$)''. Next, in time step $t_0$, the embodied agent needs to select a predefined language parsing tool to interpret the input question. This tool mainly utilizes LLMs to map natural language instructions onto the hierarchical scene graph, resulting in a logical path corresponding to the instructions. This mapping method effectively unifies natural language instructions within the same domain, allowing complex problems to be parsed into standardized patterns within the hierarchical scene graph. The question in Fig. \ref{framework} can be represented by the pattern ``$V_2 \rightarrow V_3 \rightarrow V_4 (A) \rightarrow A$''.

Next, based on the results of intent parsing, the embodied agent selects either the Rule-based Plan or the LLM-based Plan tool for task planning. The Rule-based Plan tool, combined with the hierarchical scene graph and standardized patterns, assists the agent in navigating to the optimal observation position. According to human observation habits, when observing objects ($e.g.$, sofas, potted plants), one typically observes from the upper level. For the attribute of the objects, if the task involves observing attributes that can be perceived remotely ($e.g.$, color), one typically observes from the current level. However, for the attributes required close-range perception ($e.g.$, state, material), one generally needs to move directly in front of the small object. For example, to answer the question in Fig. \ref{framework}, the agent's optimal observation point is at the book in the small object level. The agent needs to analyze the current position using visual information captured by the Observation tool and gradually complete the navigation task. The agent then calls the LLM-based Plan tool, combined with the Observation tool, to understand the visual scene. This tool thinks through the decision history and visual scene to clarify the object that needs to be observed and provides the next step plan. By using these tools in combination, the embodied agent gradually performs task planning and execution through continuous interaction with the visual environment, effectively completing complex human instructions. 

Additionally, we introduce ECQA, a more complex embodied question answering dataset based on the Habitat-Matterport 3D Research Dataset (HM3D)\cite{ramakrishnan2021habitat} and real-world scenarios. This dataset enriches the EQA dataset with more template-based questions
and adds complex multi-step questions, inspired by real-world human-robot interactions.

Our main contributions are summarized as follows:
\begin{itemize}
\item We propose an embodied agent that is enabled to tackle complex questions by the visual environment-interactive planning.
\item We create a structured semantic space to handle the process of multi-turn interaction.
\item We contribute ECQA, a new EmbodiedQA dataset that is more complex and closer to human communication.
\item Experiments have shown that our framework can effectively handle complex tasks and is efficient in the real world.
\end{itemize}

\section{Related work}

\subsection{Embodied Question Answering}
The task of Embodied Question Answering (EmbodiedQA) was first introduced by Das et al. \cite{das2018embodied} in 2018 and has rapidly emerged as a prominent research area in the fields of artificial intelligence and robotics over the past few years\cite{das2018neural}\cite{duan2022survey}\cite{luo2019segeqa}\cite{gordon2018iqa}\cite{tan2020multi}\cite{wijmans2019embodied}. The task requires the robot to follow human instructions for navigation in the environment and then provide answers based on visual perception. Some researches mainly address the navigation task \cite{zhan2024enhancing}\cite{wang2023res}\cite{chen2023think}. Zhan et al. \cite{zhan2024enhancing} focuses on room types in indoor environments to achieve more accurate navigation. Wang et al. \cite{wang2023res} leverages sequential visual information to predict descriptive instructions, aiding object-level navigation. Chen et al. \cite{chen2023think} improves navigation performance by predicting the navigation state for the next few time steps. Due to the limited generalization capabilities of state-of-the-art models of the time, human instructions only considere template-based questions. These questions typically have relatively clear intentions, making tasks easier to handle. 

With the rise of foundation models, the EmbodiedQA task has seen significant development and enhancement\cite{sakamoto2024map}\cite{patel2024embodied}. Unlike previous template-based questions, \cite{majumdar2024openeqa} and \cite{ren2024explore} propose the open-vocabulary datasets, where the question possibly requires semantic reasoning. They usually have clear and straightforward intentions, making task understanding relatively simple.

\subsection{Visual Scene Representation in Embodied Intelligence}
The research on visual scene representation aims to enable intelligent agents to understand human-level scenes\cite{bae2022survey}. Effective representations for visual environments play a crucial role for intelligent agents to gather and store information from the environments. \cite{rosinol20203d} proposes a layered and hierarchical graph that includes 5 layers to describe an indoor environment:  (i) Metric-Semantic Mesh, (ii) Objects and Agents, (iii) Places and Structures, (iv) Rooms, and (v) Building. On this basis, some research considers scene graphs from the first perspective of robots, focusing on \textit{local} information. \cite{ravichandran2022hierarchical}\cite{han2022scene}\cite{hu2023agent} focus on scene information from the current perspective. GRID\cite{ni2023grid} separates the robot from the scene graph to create a robot graph, paying more attention to the behavior of the robot. Other research considers scene graphs from the perspective of overall environment, focusing on \textit{global} information. SayPlan \cite{rana2023sayplan} pre-constructs a complete scene graph of the indoor environment to assist robots in visual language navigation. The positions, attributes, and other information of all nodes are provided in advance. SayNav \cite{rajvanshi2024saynav} incrementally builds and expands a 3D scene graph while exploring unknown environments, thereby forming a fine-grained cognition of the surroundings. 

In this paper, we adopt the global perspective to construct scene graphs. Differently, we propose that the embodied agent only needs to have prior knowledge of the steady-state objects, without transient information such as small objects.

\begin{figure}[!t]
    \centering
    \includegraphics[width=0.45\textwidth]{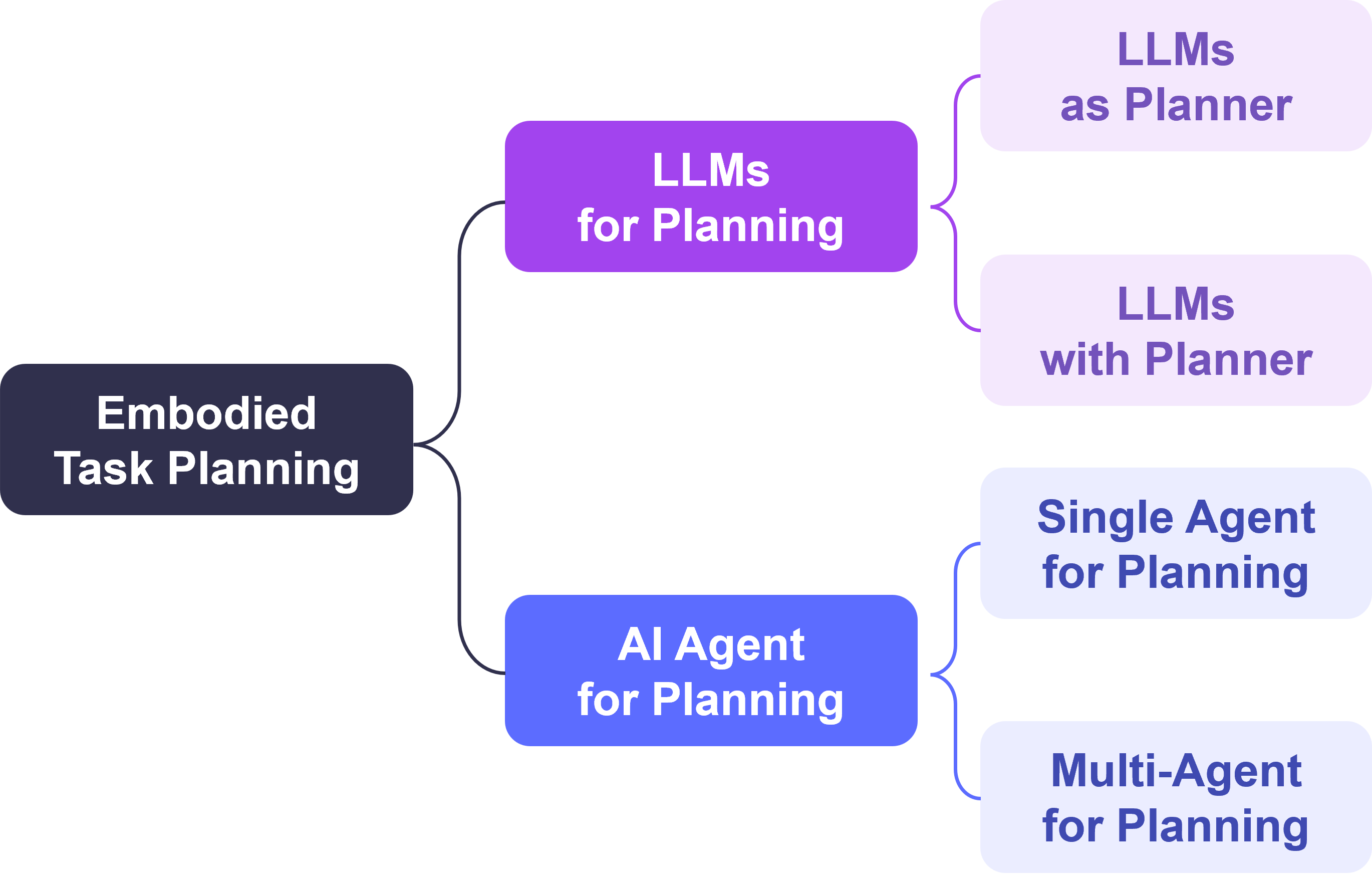}
    % where an .eps filename suffix will be assumed under latex, 
    % and a .pdf suffix will be assumed for pdflatex; or what has been declared
    % via \DeclareGraphicsExtensions.
    \caption{A taxonomy of task planning methods in the era of foundation models with the related work from this section referenced.}
    \label{Fig.2}
\end{figure}

\subsection{Embodied Task Planning}
Embodied task planning involves the decomposition of complex tasks into a sequential set of executable steps for robot execution. An overview of the different task planning approaches is shown in Fig. \ref{Fig.2}.

\subsubsection{LLMs for Planning}
LLMs-based task planning can be divided into two categories: \textit{LLMs as Planner} and \textit{LLMs with planner}. The former leverages LLMs' ability to decouple complex tasks, using them as planners to generate action sequences for embodied agents. Some research provides the solution steps for tasks in one go\cite{huang2022language}\cite{wei2022chain} by LLMs, which is suitable for scenarios where task steps are clear and have minimal dependencies. Other approaches focus on generating one action at a time, allowing for feedback from the execution to generate next action\cite{barmann2023incremental}\cite{huang2022inner}\cite{liang2023code}\cite{wu2023tidybot}\cite{singh2023progprompt}. The latter approach posits that, while LLMs excel in language processing, they lack the core reasoning abilities required for task planning. Therefore, this method primarily utilizes LLMs to translate natural language instructions into PDDL\cite{liu2023llm+}\cite{chen2024autotamp}\cite{birr2024autogpt+}, thereby assisting embodied agents in task execution. These methods overly rely on large models and are susceptible to the influence of hallucination. 

\subsubsection{AI agent for Planning}
The inherent generative capabilities of LLMs may not be sufficient to handle complex tasks. Therefore, recent research has introduced the concept of AI Agents to improve task execution accuracy. This concept mimics the thinking patterns of the human brain, endowing the agent with abilities such as perception, decision-making, action, memory, and tool use. Moreover, AI Agents can self-optimize and adjust based on execution results, continually enhancing their task performance in dynamic environments. ReAct\cite{yao2022react} proposes an agent to give a thought before performing an action.  LATS\cite{zhou2023language} synergizes planning, acting, and reasoning by using trees. RAISE\cite{liu2024llm} builds upon the ReAct by incorporating a dual-component memory mechanism, corresponding to human long-term and short-term memory. Reflexion\cite{shinn2024reflexion} generates self-reflections through linguistic feedback to reduce LLMs hallucination. AUTOGPT + P\cite{birr2024autogpt+} addresses the inherent limitations of LLMs in reasoning capabilities by combining classical models, such as object detection and Object Affordance Mapping. As task complexity continues to increase, researchers have started employing multi-agent approaches to solve problems\cite{agashe2023evaluating}\cite{chen2023agentverse}\cite{guo2024embodied}, which leverages the collaboration of multi-agent, distributing different parts of the task among them to enhance overall task execution efficiency and effectiveness.

Inspired by AI Agent, our work proposes an embodied agent. The agent continually optimizes and adjusts its behavior by interacting with the environment, and formulates the next action plan by using different tools.

\begin{figure}[!t]
    \centering
    \includegraphics[width=0.48\textwidth]{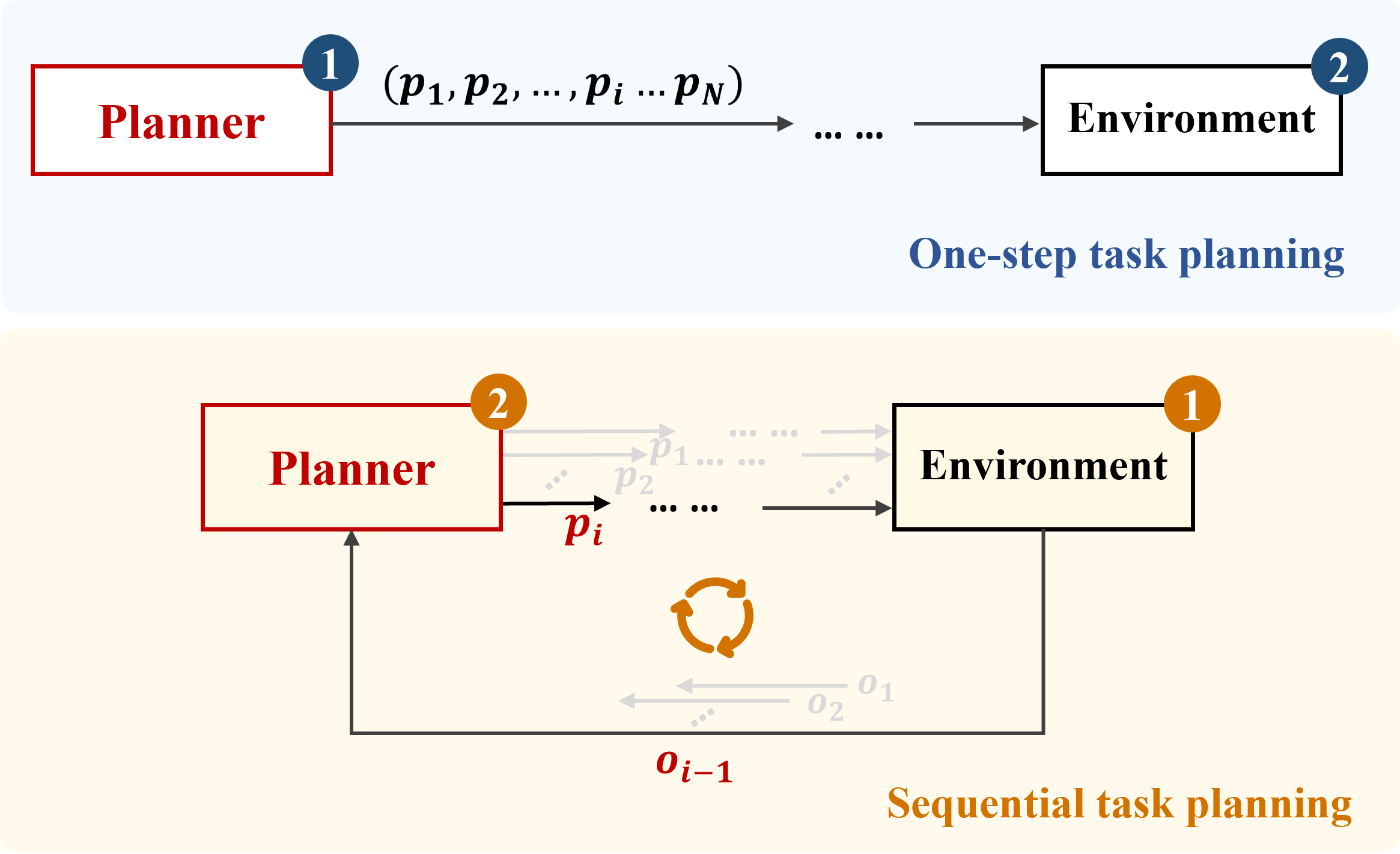}
    % where an .eps filename suffix will be assumed under latex, 
    % and a .pdf suffix will be assumed for pdflatex; or what has been declared
    % via \DeclareGraphicsExtensions.
    \caption{A comparison between one-step and sequential (ours) task planning at $i$ time step.}
    \label{fig.3}
\end{figure}

%\IEEEpubidadjcol

\section{Methodology}

\subsection{Task Definition}
Our goal is to address the challenges of complex task planning for embodied agents (such as mobile robots) in indoor environments based on natural language instructions. This requires the robot to decompose complex commands, understand visual scenes, and generate task plans that involve navigation within the environment and visual question answering. 

The goal of this task can be described as $G=f\left(Q,E\right)$. The input is two-fold: 1) a high-level natural language question $Q$ and 2) a specific indoor environment $E$ that the robot needs to explore. Firstly, $Q$ should be decomposed into $\left\{q_1,q_2,\ldots,q_n\right\}$. The intermediate results are the sequence of task plans $P=\left\{p_1,p_2,\ldots,p_n\right\}$ according to the observation feedback $O=\left\{o_0,o_1,\ldots,o_n\right\}$ during every exploration, which has shown in Fig. \ref{fig.3}. At time step $i$, the plan $p_i$ should be deduced by the current sub-question $q_i$ and previous context following the policy $\pi\left(p_i |h_i\right)$, where $h_i=\left(o_0,q_1,p_1,o_1,\ldots,p_{i-1},o_{i-1},q_i\right)$ is the history context to the robot. The final output is correct answer $G$ for $Q$.

\begin{figure}
	\centering
	\includegraphics[width=0.96\linewidth]{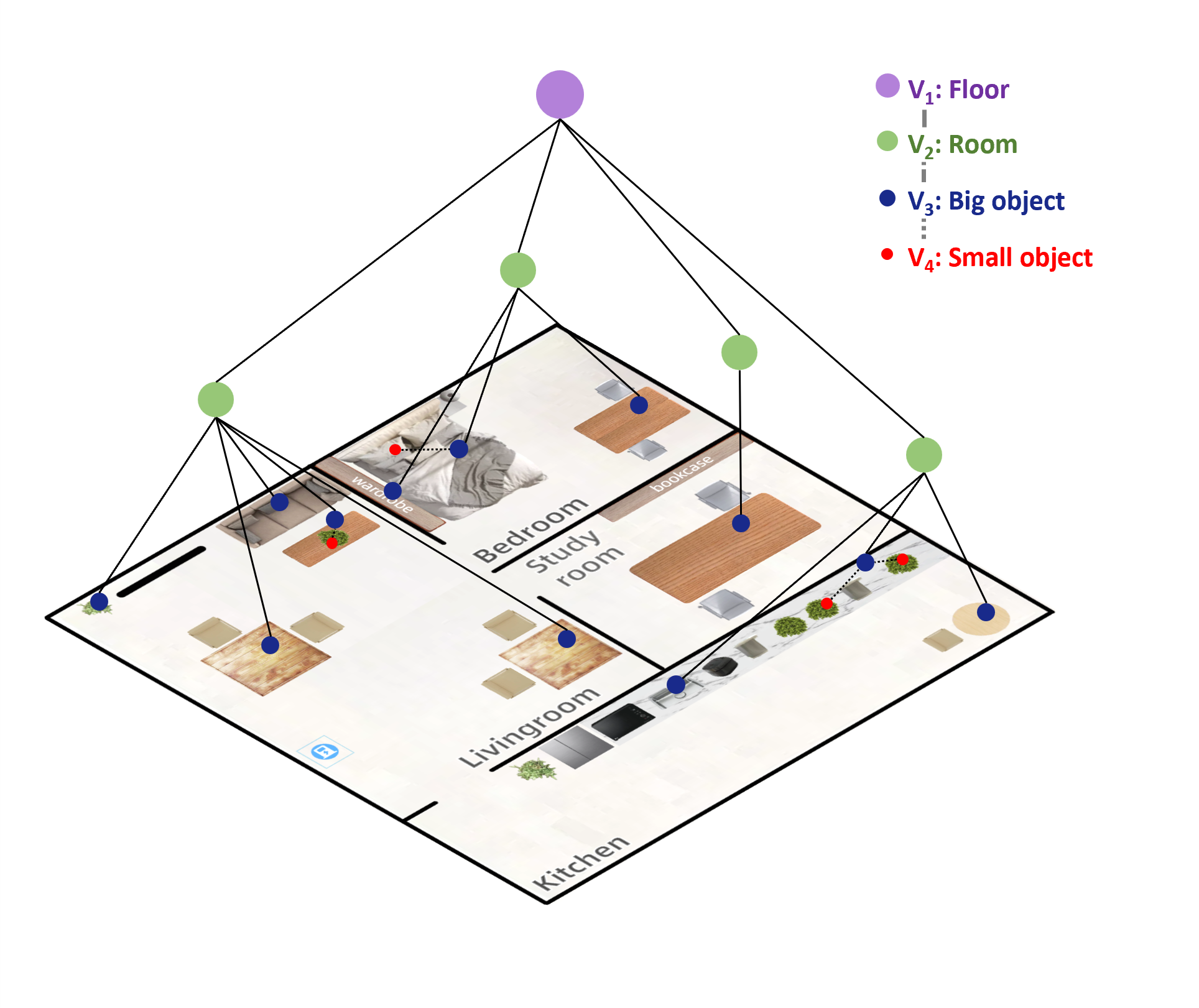}
	\caption{An example of the hierarchical structure of an indoor scene graph. The scene graph consists of four levels: floor, room, big object, and small object.}
	\label{fig:scene-graph}
\end{figure}

Here, $Q$ represents language modality information, while observation feedback $O$ is visual modality information. Analyzing and effectively integrating these two modalities is crucial for generating each plan. To achieve this, we propose a structured semantic space $S$, which is organized hierarchically. Information from different modalities can be mapped into space $S$ according to the pre-defined hierarchical structure and interact at each level, enabling more precise planning. Considering the specific scenario, the structured semantic space in Fig. \ref{framework} is defined with a four-level structure, $S = \left\{S_1, S_2, S_3, S_4\right\}$. $S_i = \left\{V_i, A\right\}$, where $V_i$ represents the set of nodes at each level, and $A$ represents the set of attributes. The input natural language $Q$ can be mapped into the space, forming a chained structure: $V_2 \rightarrow V_3 \rightarrow \left\{V_4, A\right\} \rightarrow \left\{V_4, A\right\}$. That is, $Q = \left\{q_1, q_2, q_3, q_4\right\}$, where $q_1$ belongs to $V_2$, $q_2$ belongs to $V_3$, $q_3$ belongs to $\left\{V_4, A\right\}$, and $q_4$ belongs to $\left\{V_4, A\right\}$. At time $t_1$, the language modality information is $q_1$ belonging to $V_2$, and the observed visual feedback $o_1$ should also be mapped to the $S_2$ layer in space $S$. Through intra-layer interaction, it can be determined whether the current action aligns with expectations.

\begin{table*}[t]
\renewcommand{\arraystretch}{1.5} % 调整行间距，1.15 与图中的间距相近
% if using array.sty, it might be a good idea to tweak the value of
% \extrarowheight as needed to properly center the text within the cells
\caption{Examples of Standard Pattern in EQAv1 Dataset}
\label{Intention Parsing}
\centering
% Some packages, such as MDW tools, offer better commands for making tables
% than the plain LaTeX2e tabular which is used here.
\begin{tabular}{ll}
\hline
\textbf{Natural language instructions} & \textbf{Standard Pattern} \\
\hline
What room is the $\left\langle OBJ\right\rangle$ located in? & $V_3\rightarrow V_2$ or $V_4\rightarrow V_2$ \\
\hline
What color is the $\left\langle OBJ\right\rangle$? & $V_3\rightarrow A$ or $V_4\rightarrow A$ \\
\hline
What color is the $\left\langle OBJ\right\rangle$ in the $\left\langle ROOM\right\rangle$? & $V_2\rightarrow V_3\rightarrow A$ or $V_2\rightarrow V_4\rightarrow A$ \\
\hline
What is $\left\langle on/above/below/next-to \right\rangle$ the $\left\langle OBJ\right\rangle$ in the $\left\langle ROOM\right\rangle$? & $V_2\rightarrow V_3\rightarrow V_3$ or $V_2\rightarrow V_3\rightarrow V_4$ \\
\hline
\end{tabular}
\end{table*}

\subsection{Indoor Visual Scene Representation}

\subsubsection{The Definition of Scene Graph}
Scene graphs (SG) is a structured and hierarchical way to represent the world. It provides a structured semantic space, serving as a bridge for the interaction and integration of information from different modalities. Conventional SG representation typically abstracts visual environments by utilizing discrete labels or semantic features to represent vertices and their relationships. These SGs are trained within the language domain, referred to as language SGs. Their vertices only imply intrinsic property such as object affordances, without visible attributes like object states present in the environment. Therefore, to gather visual features for answering questions, agents must navigate indoor scenes guided by a visual scene graph concluding amply attribute information. 

A language SG is denoted as $G_l = (V_l,E_l,A)$, where $V_l$ represents the set of vertices and $E_l$ connects different vertices, representing various relationships. $A$ signifies the set of visible attributes that are independent of vertices. A visual SG can be denoted as $G_v = (V_v,E_v)$, where $V_v$ stands for graph nodes in a specific scene and $E_v$ exemplifies actual orientation through weighted edges. The connection between the two modal SGs is mainly reflected in the nodes, which can be described as $V_v=\left\{A^{'}\wedge V_l | A^{'} \subseteq A\right\}$. The exploration of the robot in the environment relies on a visual scene graph, which essentially provides a structured semantic space for the integration of language expression and visual perception.

\subsubsection{The Structure of Indoor Visual Scene Graph}
The scene graph is organized in a hierarchy way with four primary layers: floor layer, room layer, big object layer, and small object layer, which has shown in Fig. \ref{fig:scene-graph}. The top level represents the floor, with each floor extending to multiple rooms such as bedrooms, living rooms, studies, and kitchens. Within each room, big objects and small objects can be found. Big objects are typically fixed furnishings, while small objects are often movable and typically associated with big objects. For instance, in a living room, big objects include a coffee table, a sofa, and 2 desks, while small objects like the potted plant is placed on the coffee table. The nodes representing big objects in the scene graph encode detailed positional information. However, additional attributes, such as state, purpose, color, or weight, are not pre-stored and must be sensed by the robot during task execution. Notably, to enhance the robot's adaptability to dynamic environments, no information about small objects is pre-defined, requiring the robot to infer it autonomously through observation. This hierarchical structure is also highly extensible and can be adapted to larger environments, such as buildings or campuses.

According to the hierarchical definition of visual scene graph, all vertices $V$ can described as  $V_1  \cup V_2 \cup V_3 \cup V_4$, with each $V_k$ signifying the set of vertices at the layer $k$. All edges can be represented as $E= E_{12}\cup E_{23}\cup E_{34}\cup E_{22}\cup E_{33}\cup E_{44}$, where $E_{ij}=\left\{V_i\rightarrow V_j |i \neq j\right\}$ means abstract relations between adjacent layers and $E_{ij}=\left\{V_i\rightarrow V_j |i =j\right\}$ means the spatial relation within the same layer. Upon entering a new environment, embodied agent must perform a comprehensive survey to construct an initial visual scene graph. This graph should encompass all objects, attributes, and their spatial coordinates within the floor level, room level, and large object level. The agent is required to dynamically update the scene graph through active perception and exploration, guided by human instructions.

\begin{algorithm}
	\renewcommand{\arraystretch}{1.5} % Adjust row height
	\caption{The combination of Rule-based Plan and LLM-based Plan tool}
	\label{algorithm_rule_LLM}
	\begin{algorithmic}[1]
	  \State $Q \gets$ Question
	  \State $G_v \gets {V_1, V_2, V_3, V_4}$ \Comment{Visual hierarchical scene graph}
	  \State $Pattern \gets$ Output of language parsing
	  \State $t \gets 0$ \Comment{Time step}		
  	  \State $k \gets 0$ \Comment{Subgoal index}
  	  \State $o \gets$ None \Comment{For Observation tool}
	  \State $target \gets Pattern[-1]$ 
	  \State $Flag \gets 0$
	  \While{$k < \text{len}(Pattern)$}
		\If{$k = \text{len}(Pattern)-1$}
			\If{$target$ is $obj$ in $V_i$}
				\State $p_t \gets$ Move to the parent node in $V_{i-1}$ \Comment{Rule}
				\State $o \gets  \text{LLM}(p_t, Pattern, G_v)$	 \Comment{LLM}
			\ElsIf{$target$ is $attr$ $A$ of $obj$ in $V_i$}
				\State $\text{attr\_cate} \gets \text{LLM}(A, obj, G_v)$ \Comment{LLM}
				\If{\text{attr\_cate} is close-range perception}
					\State $p_t \gets$ Move to the object in $V_i$	\Comment{Rule}			
				\Else
					\State $p_t \gets$ Move to the parent node in $V_{i-1}$ \Comment{Rule}
				\EndIf
				\State $o \gets \text{LLM}(p_t, Pattern, G_v)$ \Comment{LLM}
			\EndIf		
		\Else
			\State $p_t \gets$ Move to $Pattern[k]$ \Comment{Rule}
			\State $o \gets \text{LLM}(p_t, Pattern, G_v)$ \Comment{LLM}
		\EndIf	
		\State $k \gets k+1$			
		\State $t \gets t+1$				
	  \EndWhile	
	\end{algorithmic}
\end{algorithm}

\begin{algorithm}
\renewcommand{\arraystretch}{1.5} % Adjust row height
\caption{Visual Environment-Interactive Task Planning}
\label{algorithm_overall}
\begin{algorithmic}[1]
\State $Q \gets$ Question
\State $G_v \gets$ Visual hierarchical scene graph
\State $Tools \gets$ List of all tools
\State $G \gets$ The final answer
% \State $H \gets$ History context
\State $t \gets 0$ \Comment{Time step}
\State $k \gets 0$ \Comment{Subgoal index}
\State $T_{language\mbox{-}parsing} \gets Tool\mbox{-}Selection(Tools)$ 
\State $Pattern \gets T_{language\mbox{-}parsing}(Q)$
\State $q \gets Pattern[k]$ \Comment{First subgoal}
\State $Flag \gets 0$
\While{$k < \text{len}(Pattern)$}
	\If{The target in $Pattern$ is the attribute $A$}
		\State $T_{LLM\mbox{-}based Plan} \gets Tool\mbox{-}Selection(Tools)$ 
		\State $A_{Type} \gets T_{LLM\mbox{-}based Plan}(A)$ \Comment{Attribute type}
	\ElsIf{The target in $Pattern$ is the object}
		\State $A_{Type} \gets \text{None}$	
	\EndIf
	\State $T_{Rule\mbox{-}based Plan} \gets Tool\mbox{-}Selection(Tools)$ 
	\State $p_t \gets T_{Rule\mbox{-}based Plan}(q, A_{Type})$
	\State execute $p_t$
	\State $T_{Observation} \gets Tool\mbox{-}Selection(Tools)$ 
	\State $o_t \gets T_{observation}(p_t)$ 
	\If{$check\mbox{-}feedback(o_t, p_t) = True$}  \Comment{$p_t$ is completed}
		\State $Flag \gets 1$
	\ElsIf{$check\mbox{-}feedback(o_t, p_t) = False$} \Comment{$p_t$ is not completed}
		\State $Flag \gets 0$        		
	\EndIf
	\If{$Flag \gets 1$}
		\State $k \gets k + 1$
		\State $q \gets Pattern[k]$ \Comment{Get next subgoal}
		\State $t \gets t + 1$
		\State $Flag \gets 0$
	\EndIf
\EndWhile
\State $G \gets o_t$
\end{algorithmic}
\end{algorithm}

\subsection{Approach}
We propose an embodied agent framework that can gradually formulate plans through continuous interaction with the visual environment. This framework is grounded in a structured semantic space, which enable sequential ($i.e.$, step-by-step) planning. As illustrated in Fig. \ref{fig.3}, unlike the one-step planning, our method requires multi-turn feedback (short-term planning) from the environment. Such a mechanism allows the system to promptly adjust each step of the plan based on environmental observations, facilitating efficient error diagnosis or debugging. The process is essentially a ``Observation-Planning-Action'' cycle. Within the cycle, we design a series of tools (Language Parsing, Rule-based Plan, LLM-based Plan, Observation) to collaboratively handle complex tasks. Language Parsing tool parses the user's natural language instructions. LLM-based Plan tool determines the type of instruction and identifies the content to be observed in each step. Rule-based Plan tool infers the robot's movements based on pre-defined rules. Observation tool gathers visual information from the environment. The performance of our framework results from the collaborative effects of all tools.

\subsubsection{Language Parsing}
Language is the primary tool for human communication, facilitating the transmission of information and emotions through speech, text, and other forms. In embodied complex-question answering tasks, robots engaged in human-robot interaction often encounter questions that require intricate reasoning. Clarifying human intent from instructions beforehand can effectively constrain the robot's reasoning process, thereby enabling it to respond more accurately to human needs.

LLMs exhibit relative weaknesses in reasoning, which is crucial for accurate task planning. Therefore, we designs a language parsing tool that can project natural language instructions into the structured semantic space. This tool converts complex questions into standardized patterns, thereby supporting subsequent task planning and execution. Table \ref{Intention Parsing} presents examples of standard patterns in EQAv1\cite{das2018embodied} dataset. For the first type of question, ``What room is the $\left\langle OBJ\right\rangle$ located in?'', the pattern is either $(V_3\rightarrow V_2)$ or $(V_4\rightarrow V_2)$. The object mentioned in the question could be a small object or a large object, indicating that the node in the scene graph could be from either the large object layer $V_3$ or the small object layer $V_4$. The goal is to infer the corresponding node in the room layer $V_2$. After parsing the pattern, the target object that the question requires to solve and its hierarchical position in the scene graph can be obtained. This information will assist in subsequent robot navigation for path planning. Considering the feasibility and practicality, the algorithm needs to be deployed on real-world robots to achieve natural human-robot interaction. In real-world environments, under the constraints of low computational power, high reasoning speed, and human experience, the algorithm must ensure reasoning accuracy while using the smallest possible large language model. Therefore, this paper employs instruction fine-tuning to optimize a large model with 7 billion parameters. 

\begin{figure*}[htbp]
	\centerline{
		\includegraphics[width=16cm]{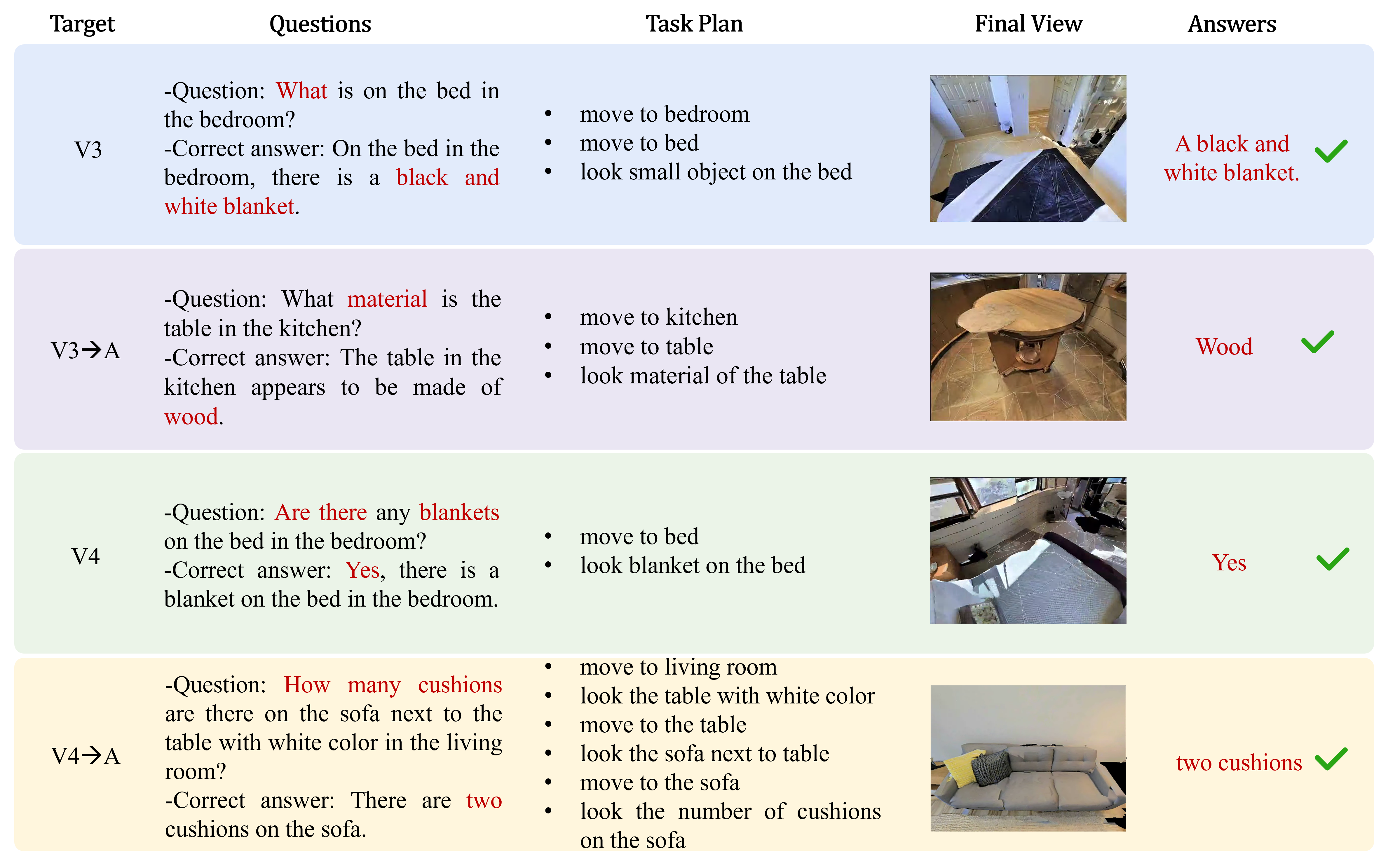} 	
	}

	\caption{Examples of our framework experiment results in simulated environments. We show the results for different types of questions and stopping steps for our method.}
	\label{example}
\end{figure*}

\subsubsection{Task Planning}
In embodied complex question answering tasks, the quality of human-robot interaction results is closely tied to the observation position of the embodied agent. Therefore, task planning needs to ensure that the robot moves to the appropriate location for observation. This paper proposes two task planning tools: Rule-based Plan tool and LLM-based Plan tool. The Rule-based Plan tool primarily determines the optimal action path for the robot based on the results of language parsing and its current position, while the LLM-based Plan tool provides guidance on what the robot needs to observe after it completes the movement, ensuring successful task execution.

\textbf{Rule-based Plan} tool draws inspiration from human habits of observing the world to plan the robot's action path. As illustrated in Fig. \ref{framework}, for a question like ``What is the title of the book open on the table in the living room?'', simply moving to the table's location may make it difficult to identify fine-grained content like ``the title of the book'' within the current field of view. To address this challenge, our Rule-based Plan tool uses the output from the Language Parsing tool to formulate an action plan. The specific rules are as follows:

\begin{itemize}
\item[-] If the target is an object that belongs to the $V_4$ level (small object layer), the embodied agent needs to move to the upper level $V_3$ (big object layer) for observation.
\item[-] If the target is an object that belongs to the $V_3$ level (big object layer), the embodied agent needs to move to the upper level $V_2$ (room layer) for observation.
\item[-] If the target is an attribute of an object in the $V_3$ level (big object layer), and the attribute is something like color, which can be perceived remotely, the embodied agent only needs to move to the $V_2$ level (room layer) for observation, as this level provides a clear view of external characteristics. When the attribute is something like material, which requires close-range perception, the agent needs to move to the $V_3$ level (big object layer) to observe the object up close.
\item[-] If the target is the attribute of an object that belongs to the $V_3$ level (big object layer), the embodied agent needs to move to the current level $V_3$ (big object layer) for observation.
If the target is an attribute of the object in the $V_4$ level (small object layer), and the attribute can be perceived remotely, the embodied agent only needs to move to the $V_3$ level (big object layer) for observation. When the attribute requires close-range perception, the agent needs to move to the $V_4$ level (small object layer) to observe the object.
\end{itemize}

Based on the aforementioned rules, the embodied agent can quickly determine the destination using the results from the Language Parsing tool. 

\textbf{LLM-based Plan} tool primarily leverages the vast commonsense knowledge of LLMs to perceive and reason about the external environment. The Rule-based Plan tool can formulate different action rules based on the characteristics of object attributes, while the determination of whether an attribute requires remote or close-range perception relies on the LLM-based Plan tool for pre-judgment. 

Moreover, the LLM-based Plan tool leverages powerful language processing capabilities of LLMs to generate the specific content that needs to be perceived based on the current step of the action plan. At the same time, using the vast commonsense knowledge of LLMs, the agent can effectively perceive and reason about the external environment. As shown in Fig.\ref{framework}, once the robot moves step by step to the open book, the original question ``What is the title of the book open on the table in the living room?'' should be simplified to ``What is the title of the book?''. This simplification helps prevent overly complex grammar from causing hallucinations during reasoning by the large model. 

Specifically, The process of collaboration between Rule-based Plan tool and LLM-based Plan is presented in Algorithm \ref{algorithm_rule_LLM}. The Rule-based Plan tool mainly determine the optimal observation position for each task based on pre-defined rules. The LLM-based Plan tool leverages the vast common-sense knowledge of LLMs, enabling it to recognize task types (such as inquiring about objects or attributes, or determining whether remote or close-range observation is required) and infer the content that needs to be observed. Two planning tools cooperate in a chain to formulate each step of the plan.

For instance, as illustrated in Fig. \ref{framework}, the natural language instruction is mapped into the structured semantic space by Language Parsing tool, forming a chain structure: $(V_2\rightarrow V_3\rightarrow V_4(A) \rightarrow V_4(A))$. The target within this instruction is an attribute $A$ of the object in $V_4$. Within this space, the Rule-based Plan tool sequentially formulates plans in a top-down manner. At time $t_1$, the plan is to move to the living room ($V_2$). At time $t_2$, the plan is to move to the coffee table ($V_3$). The LLMs-based Plan tool ascertains that the attribute necessitates close-range observation. Then, at time $t_3$, the plan involves moving to the open ($A$) book ($V_4$). Following the rule-based sequential movement to the observation position, at time $t_4$, LLMs-based Plan tool formulates the plan to "look the title ($A$) of the book ($V_4$)."

\subsubsection{Observation}
The Observation tool is primarily responsible for performing visual perception tasks. On the one hand, it must perceive the current external environment, assess whether each step of the navigation is executed correctly, and promptly provide feedback to the embodied agent to ensure the accuracy of path planning. On the other hand, it needs to integrate the perceptual requirements output by the LLM-based Plan tool to observe the target object and produce the correct answer. This process ensures that the embodied agent can accurately identify and process the necessary information in complex tasks, leading to correct decisions and responses. If a perception error occurs (i.e., the Observation tool determines that the action has failed, but the robot actually complete the action smoothly), it is necessary to combine the visual scene graph and the current position of the robot for secondary perception to confirm whether the judgment is correct.

In summary, based on the aforementioned tools, when the embodied agent receives a natural language question from the human, it first invokes the Language Parsing tool to map the question onto the visual hierarchical scene graph, breaking the question down into standardized patterns. The agent then selects the appropriate Task Planning tool and, in combination with the Observation tool, formulates the plan for the current time step. During task execution, the agent continuously observes the environment, and the observation results are used to assist in formulating the plan for the next time step, ensuring the continuity and accuracy of task execution. The detailed algorithmic flow is shown in Algorithm \ref{algorithm_overall}.

\section{Experiments}
The ultimate goal of an embodied agent is to answer questions accurately.
We now describe our experiments in detail. 1) We aim to  
introduce ECQA dataset. 2) We evaluate the results of the framework on ECQA by LLM and show its performance over other agent methods. 3) We provide an analysis to highlight the benefits of not relying on LLMs' decisions in our approach. 4) We execute the framework for our household environments, demonstrating its practicality for real-world scenarios.

\begin{table}[h!]
	\renewcommand{\arraystretch}{1.5} % Adjust row height
	\centering
	\caption{LLM-Match evaluation scores of our method and ReAct for planning with different backbones.}
	\label{Different backbones}	
	\begin{tabular}{cccccc}
		\hline
		\multirow{3}{*}{\textbf{Method}} & \multicolumn{5}{c}{\textbf{Backbone}} \\ \cline{2-6}
		& \multirow{2}{*}{\textbf{ChatGPT-4O}} & \multicolumn{4}{c}{\textbf{Qwen2.5}} \\ \cline{3-6}
		&  & \textbf{72B} & \textbf{32B} & \textbf{14B} & \textbf{7B} \\ \hline
		ReAct & 61.8 & 55.8 & 53.9 & 48.7 & 21.6 \\ 
		Ours & 65.4 & 64.4 & 61.8 & 57.9 & 46.9 \\ \hline
	\end{tabular}
\end{table}

\begin{table}[!t]
	\renewcommand{\arraystretch}{1.5} % Adjust row height
	\setlength{\arrayrulewidth}{0.1mm}
	\centering
	\caption{LLM-Match evaluation scores of our method and ReAct on template-based and multi-step questions in the ECQA dataset.}
	\label{Different question type}
	\begin{tabular}{ccc}
		\hline
		\multirow{2}{*}{\textbf{Method}} & \multicolumn{2}{c}{\textbf{Question Type}} \\ % Corrected closing brace
		\cline{2-3}
		& \textbf{Template-based question} & \textbf{Multi-step question} \\ % Corrected missing & symbol
		\hline
		ReAct & 59.1 & 52.6 \\ 
		Ours  & 66.8 & 62.0 \\
		\hline
	\end{tabular}
\end{table}

\begin{figure*}[htbp]
	\centering
	\begin{subfigure}{0.45\linewidth}
		\centering
		\includegraphics[width=\linewidth]{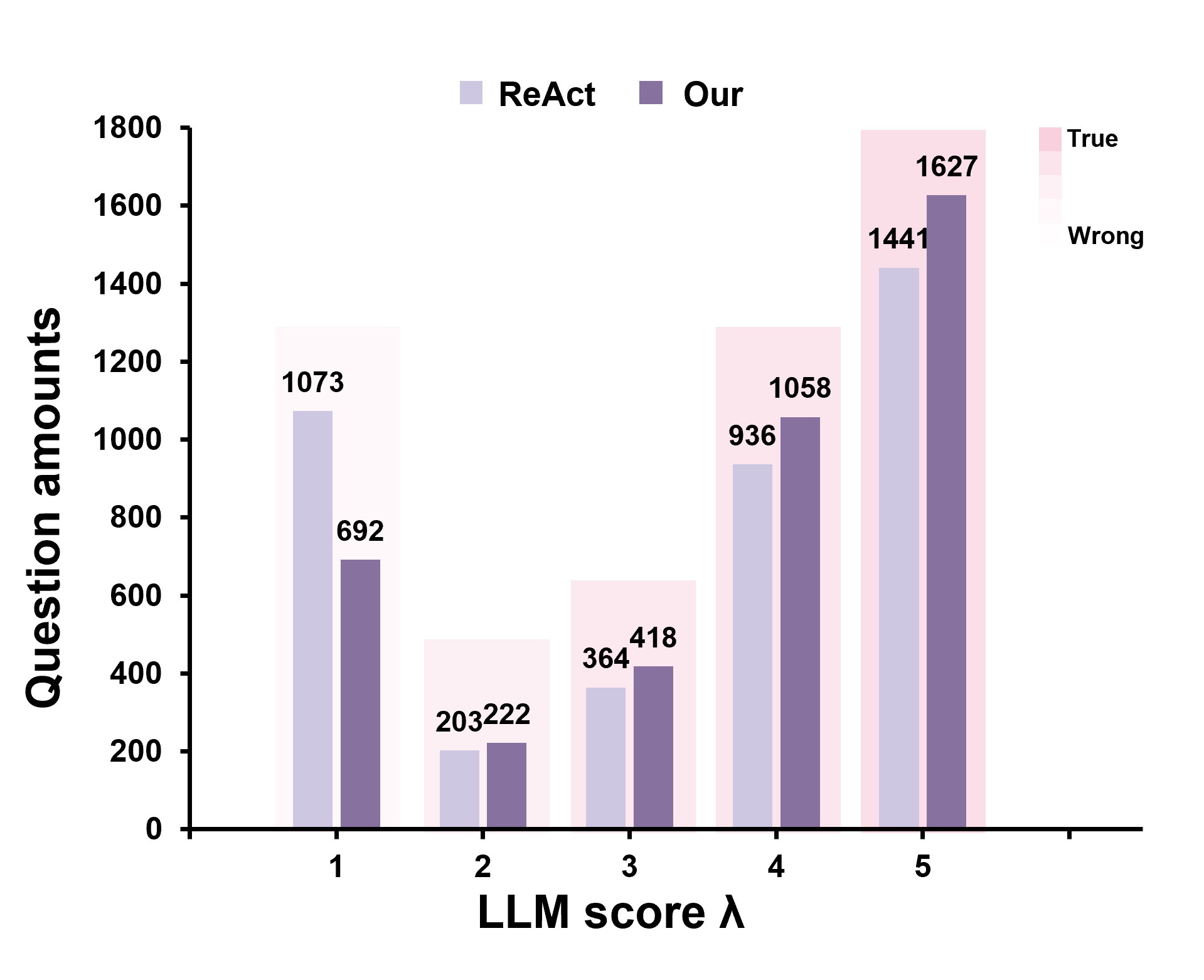}
		\caption{}
	\end{subfigure}
	\begin{subfigure}{0.45\linewidth}
		\centering
		\includegraphics[width=\linewidth]{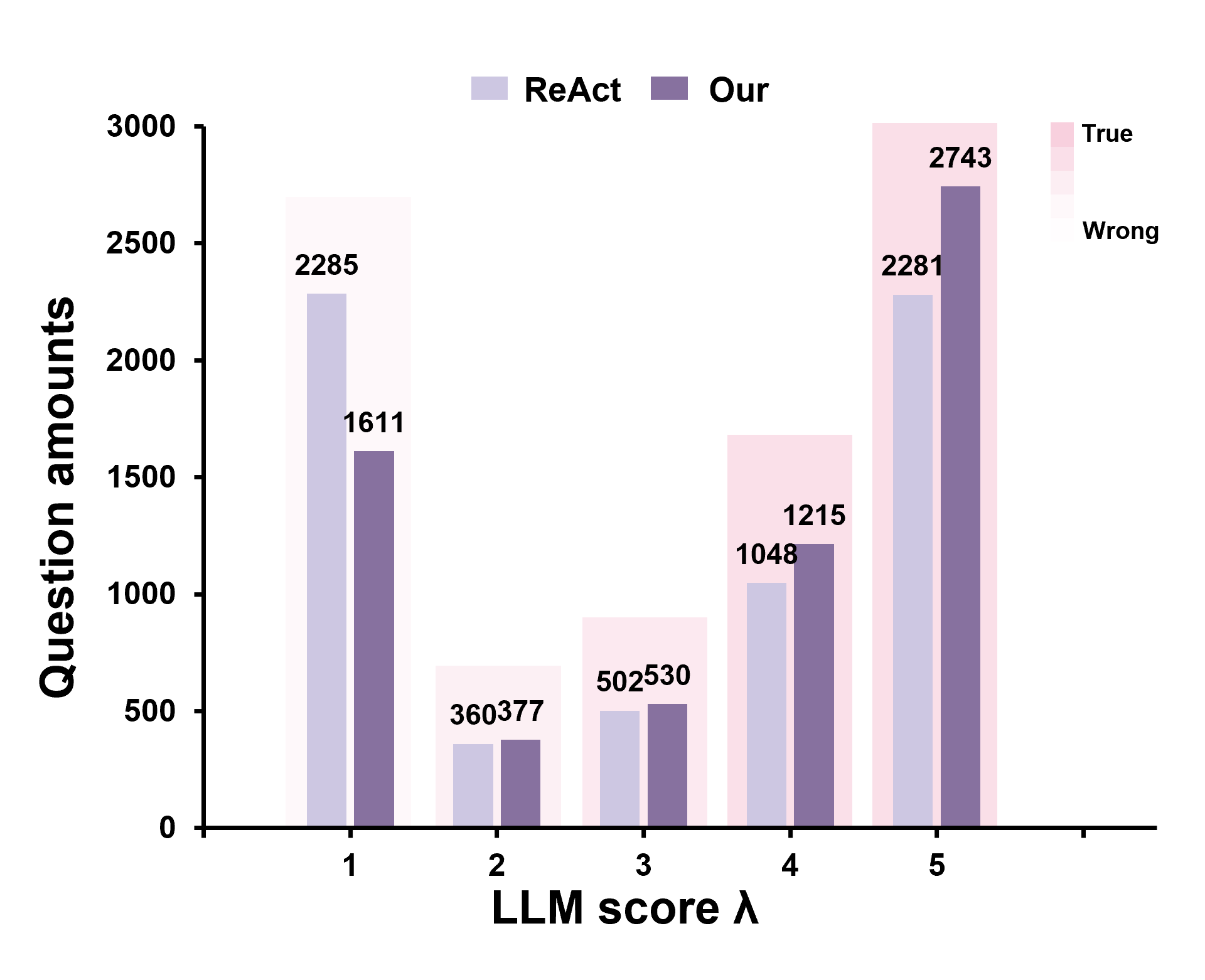}
		\caption{}
	\end{subfigure}
    \caption{LLM-Match score comparison of template-based and multi-step questions in HM3D between ReAct and our framework. (a) shows the experiment results analysis of template-based questions, (b) is the results of multi-step questions.}
	\label{fig_analysis}
\end{figure*}

\subsection{Datasets}
Previous work has primarily focused on simple questions within simulated environments, such as ``What is the color of the bed?'' or ``How many tables are there in the living room?''. These questions typically involve basic attributes of relatively large pieces of furniture and clear structure, resulting in lower complexity. Considering the complexity of human-robot communication in real world, we are now interested in applying our embodied agent framework to more diverse scenarios, where the questions might be more complex and may require more sophisticated language parsing to accurately grasp the intent of the question. Therefore, we propose ECQA dataset, which generates more questions by extending the template types of the EQA dataset, and introduces multi-step questions ($e.g.$, ``What brand is the phone held by the person sitting on the sofa in the living room?'') based on the Habitat-Matterport 3D (HM3D) simulation environment. Additionally, to enhance the diversity of questions, we also design to incorporate questions about small objects and people for real-world scenarios. Given that the simulation environment involves limited small objects and people-related content, we mainly verify these two types of questions in real environments.

The embodied robot explores different simulation environments and a self-built real indoor environment separately, storing the perceived visual information locally. Then, we utilize ChatGPT-4o to generate multi-step question-answer pairs, as well as related to small objects and people based on above information. Subsequently, we performed manual validation and removed overly simple questions($e.g.$, ``Where is the person?'').The resulting questions can be divided into four categories:

1) Template-based questions: asking about the location, quantity, material, color, and existence of objects in the HM3D environment, $e.g.$, ``What texture is the sofa in the family room?''

2) Multi-step questions: asking about complex grammar, $e.g.$, ``What is the person in a black shirt sitting on the sofa in the living room doing?''

3) Questions related to small object: asking about the number, color, state, location, and existence of small objects, $e.g.$, ``Is the laptop cover on the desk closed?''

4) Questions related to people: asking about the state, location, and existence of people, $e.g.$, ``Is the person in bed asleep?''

\subsection{LLM-Match Evaluation}
We use an LLM to evaluate the correctness of answers by embodied robot based on different task planning model for complex questions. Specifically, we adapt the evaluation protocol introduce in \cite{majumdar2024openeqa}. Given a question $q_i$, correct answer $a_i^*$, and model output answer $a_i$, the LLM should be prompted to provide a score $\lambda$,
\begin{equation}
    \lambda = LLM(a_i|a_i^*, q_i),\,with\,\lambda \in \left\{1,2,\ldots,5\right\}
\end{equation}
where $\lambda = 1$ is an wrong answer, $\lambda = 5$ is a correct answer, and others indicate the similarity of answers. Finally, the evaluation results $C$ based on LLM are represented as follows:
\begin{equation}
    C = \frac{1}{N} \sum_{i=1}^{N} \frac{\lambda_i - 1}{4} \times 100\%
\end{equation}
where $N$ is the number of questions.

\subsection{Implementation Details}
We first conduct experiments in the HM3D simulation environment. We select the classic ReAct\cite{yao2022react} from the methods in a once-for-all manner and compare it with our method, which adopts a sequential manner. In the experiments, we use various LLMs as backbones. Selecting an appropriate backbone is critical, as it must account for task complexity, model size, and so on. In order to enable the deployment of our framework on a robotic platform and facilitate assistance in real-world scenarios, we have considered two aspects as outlined below.

Open-source and closed-source are two types of LLMs, which are the first aspect we consider. We select Qwen2.5-72B (953.84 TFLOPS) and ChatGPT-4O as our backbones respectively, with strongest logical reasoning and language understanding capabilities according to the rankings from OpenCompass. Closed-source models typically rely on API calls and cannot be deployed locally to support flexible development. To enhance practicality in the real world, we are prone to adopt open-source models.

Parameter size of the open-source model is the second aspect we consider, which will affect the deployment of the model on physical robots. We select smaller models as our backbones, namely Qwen2.5-32B (427 TFLOPS), Qwen2.5-14B (189.78 TFLOPS), and Qwen2.5-7B (91.97 TFLOPS). The reason is that the models with larger parameter sizes typically have higher hardware requirements and significantly increase memory and computational demands. We prefer to select smaller models while ensuring performance. Therefore, we progressively reduce the parameters of LLMs to compare their performance.

Then, we deploy the framework on robots in real-world scenarios. The robot is equipped with the edge terminal NVIDIA Jetson AGX Orin (64GB memory). To facilitate deployment on the Orin platform, we use Qwen2.5-14B for planning according to the result of the simulation environment. In the process of actual observation, object occlusion is an inevitable issue. To address this, we employ Capsule Network methods \cite{10384753}\cite{9334445}, which are effective at capturing spatial relationships between objects, thereby reducing the information loss caused by occlusion.

Additionally, we pre-built the first three layers of the visual hierarchical scene graph using ground truth and used them as prior knowledge to assist task planning of the robot. Due to the mobility and state variability of small objects, the fourth layer of the visual hierarchical scene graph is not pre-defined.

\subsection{Experimental Results in Simulated Environments}
In the simulation environment, we have conducted two comparative experiments. The first experiment used LLMs with different parameter sizes as backbones to compare the performance of our method and ReAct. The second experiment keeps the backbone consistent and compares the performance of our method and ReAct on template-based and multi-step questions in the ECQA dataset.

Firstly, Tab. \ref{Different backbones} reports the LLM-Match scores of using LLMs with different types and parameter sizes as backbones on the ECQA dataset. Based on these results, we conduct the following analysis:

1) ChatGPT-4O: According to OpenAI's performance tests on ChatGPT-4O, the model demonstrates superior capabilities in reasoning, question answering, and other tasks. Experimental results further validate this, showing that the model achieved the best results in both our method and ReAct.

2) Qwen2.5-72B: Compared to closed-source models, open-source models still exhibit a noticeable gap in reasoning capabilities. Experimental results show that ReAct is directly affected by the reduction in parameters, with its performance dropping to 55.8\%. In contrast, our approach remains relatively stable, achieving 64.4\% score.

3) Qwen2.5-32B: The experimental results show that LLM-Match score of ReAct drops from 55.8\% to 53.9\%, while our method decreases from 64.4\% to 61.8\%. Both methods exhibit a similar rate of decline.

4) Qwen2.5-14B: The performance of ReAct drops from 53.9\% to 48.7\%, while our method decreases from 61.8\% to 57.9\%. It is evident that reducing the parameter size to 14B has a more significant impact on ReAct.

5) Qwen2.5-7B: When the parameter size reduces to 7B, the reasoning capabilities of LLMs experience a significant decline. Experimental results show that both methods lead to a sharp decrease in performance.

Based on these results, we draw the following conclusions. For the comparison between open-source and closed-source models, ChatGPT-4O is only slightly better than Qwen2.5-72B. For the comparison of open source models with different parameter sizes, as the number of parameters decreases, the performance also declines. Notably, when the parameter size drops to 7B, the performance has sharply declined. Therefore, Qwen2.5-14B is a better choice, as it achieves strong performance with a relatively smaller parameter size.

Secondly, we compare the LLM-Match scores of our framework and ReAct on template-based and multi-step questions within the ECQA dataset, which has shown in Tab. \ref{Different question type}. The demonstration examples on our framework are shown in Fig. \ref{example}. Detailed experimental results are shown in Fig.\ref{fig_analysis}, which illustrates the distribution of the LLM-Match score for our method and ReAct across two types of questions. Based on the results, we share some observations and comments.

\begin{figure*}[htbp]
	\centerline{
		\includegraphics[width=16cm]{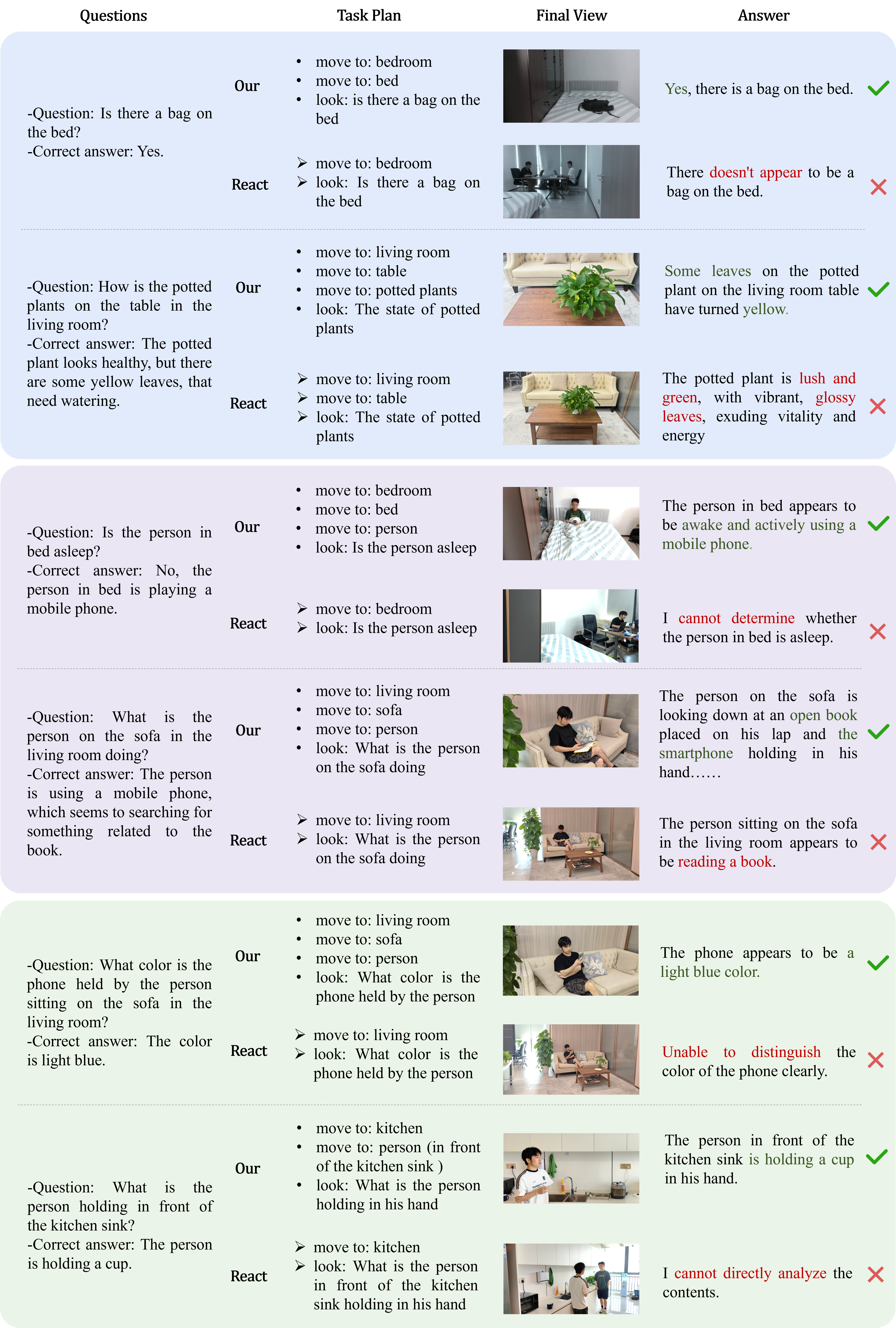} 	
	}

	\caption{Examples of our framework experiment results in real environment. We show experimental results examples for three types of questions: small objects, people, and multi-step.}
	\label{fig_real-world_demonstration}
\end{figure*}

\begin{figure*}[htbp]
	\centering
	\includegraphics[width=16cm]{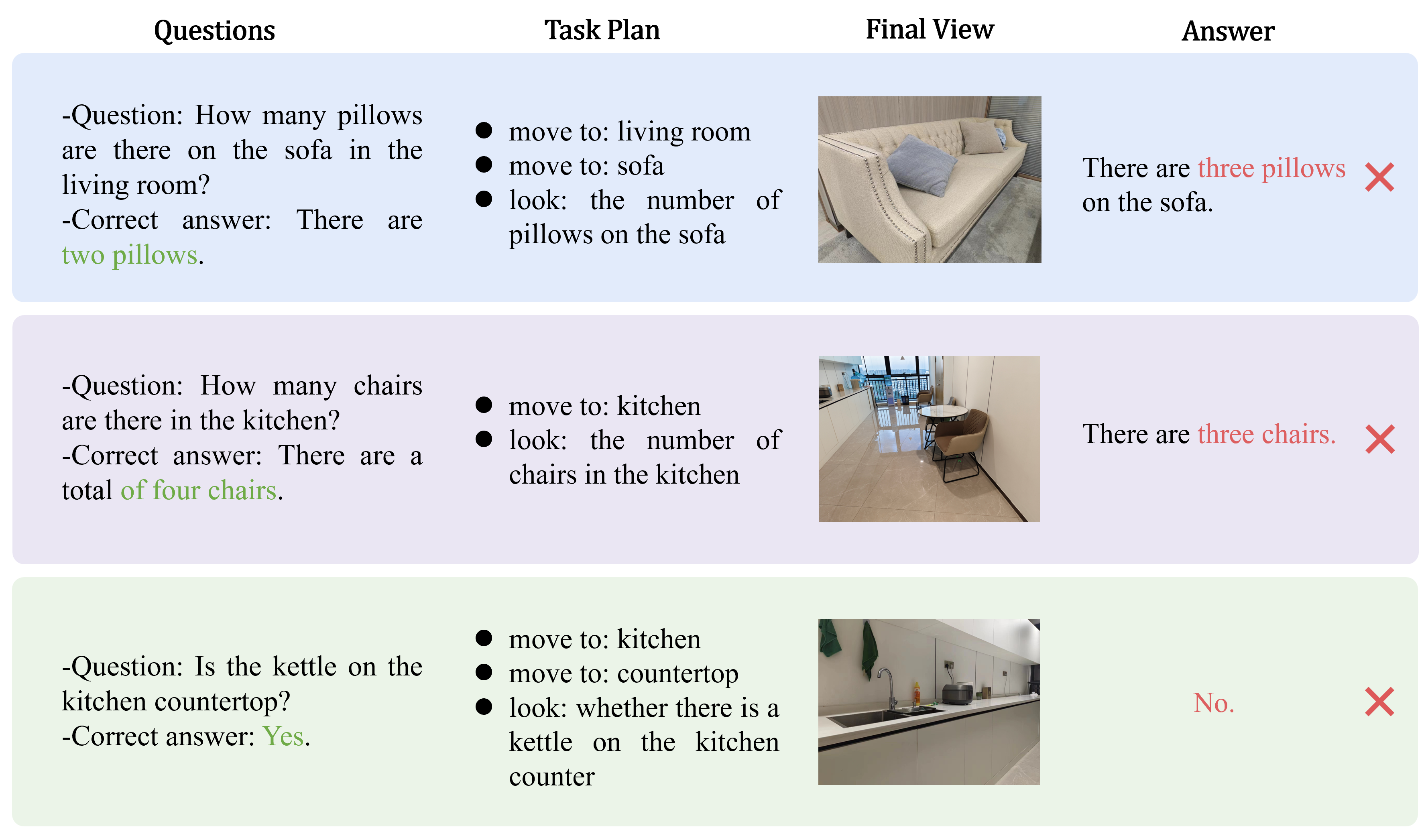}
	\caption{The failure cases of our method in real-world.}
	\label{failure case}
\end{figure*}

1) Tab. \ref{Different question type} shows that our framework performs exceptionally well on both template and multi-step questions, with LLM-Match scores significantly higher than those of ReAct.

2) For template-based questions, our framework consistently outperforms the ReAct across most LLM-Match score intervals (66.8\% vs 59.1\%). Fig.\ref{fig_analysis} (a) shows that our framework is better suited to produce high-quality answers. Meanwhile, ReAct tends to generate more low-scoring matches, especially in the $\left[1,2\right)$ interval, indicating weaker performance in those cases. Take the first question of the purple box in Fig. \ref{fig_real-world_demonstration} as an example: ``Is the person in bed asleep?''. ReAct, leveraging the pre-constructed visual scene graph, identified that the bed is located in the bedroom and move to the bedroom for observation. However, from the final observed view of the robot in the figure, it is evident that due to visual occlusion the current position does not even allow confirmation of whether there is a person in bed. In contrast, our method firstly utilizes the Rule-based Plan tool to identify that the question pertains to the state attributes of a large object in the visual scene graph. Subsequently, the LLM-based Plan tool determines that this attribute requires close-range observation, prompting the robot to move closer to the person in bed for further inspection.

3) For multi-step questions, our framework is also better than ReAct (62\% vs 52.6\%). The specific results are shown in Fig.\ref{fig_analysis} (b). It can be observed that the number of wrong tasks answered by ReAct (2285) even exceeds the number of correct ones (2281). Taking the first question in the green box of Fig. \ref{fig_real-world_demonstration} as an example, the question is: "What color is the phone held by the person sitting on the sofa in the living room?". For this type of multi-step question, the ReAct method relies on the LLMs for reasoning. However, the model does not consider the specific visual environment, and simply moves the robot to the living room for observation. In contrast, our method first analyzes that the question pertains to the attribute of a small object in the visual scene graph, which requires fine-grained observation. After planning to move to the living room, the framework uses visual perception results to determine that the current position cannot capture information about the target object, and proceeds with further planning until the robot reaches the optimal observation position.

Both experiments above validate that our method outperforms ReAct. The reason is that our framework is grounded in a structured semantic space, which enable sequential ($i.e.$, step-by-step) planning. Unlike the one-step planning of ReAct, our method requires multi-turn feedback (short-term planning) from the environment. Such a mechanism allows the system to promptly adjust each step of the plan based on environmental observations, facilitating efficient error diagnosis or debugging.

\subsection{Real-World Demonstration}
We also port the framework to a real embodied robot and test under various settings, for showcasing the efficient generalization of the framework to real environments. We construct a real-world indoor home environment, as shown in the Fig.\ref{fig:scene-graph}, which includes four rooms: living room, kitchen, bedroom, and study. Each room is furnished with various large and small objects commonly used in daily life, and there are people moving normally within the environment. We validated the questions related to small objects, people, and multi-step from self-constructed ECQA dataset in real-world scenarios. The demonstration examples are shown in Fig. \ref{fig_real-world_demonstration}.

For questions, such as ``Is there a bag on the bed'', which involve small objects, comparative experiments revealed that our method successfully navigates to the front of the bed, avoiding visual blind spots. For questions about small object attributes, our approach effectively moves in front of the object for observation, preventing incomplete observations caused by occlusions. For questions concerning a person's status, our method accurately moves to the person's location to prevent missing information due to distance. In multi-step questions, our approach uses the hierarchical scene graph to progressively reason and provide a plan, successfully avoiding distractions from other elements ($e.g.$, ``a person in white clothing on the bed'').

Additionally, Fig. \ref{failure case} illustrates failure cases of our method in real-world experiments. Task the question in the blue box as an example, the robot has moved near the sofa to observe the number of cushions. The decision made by our method is correct. However, the complexity of the real-world environment poses challenges. In our visual hierarchical scene graph, the room level and large object level both encode pre-defined location information. Due to viewpoint occlusion at the defined sofa location, the robot could not fully observe all objects, resulting in an incorrect final answer. In future research, we can explore the robot's active perception capabilities. By enabling the robot to adaptively adjust its observation angle according to user instructions, it can capture more comprehensive information.

\section{Limitations}
Our framework has explored the complexity of human instructions, but in real-world applications, the instructions encountered are not only complex problems but also more everyday, oral questions. Oral questions are more flexible and random, making it more difficult to capture intent. This requires embodied robots to have a rich knowledge base that supports multi-turn conversations with humans to accurately determine their intentions. Future work could focus on real-world applications, specifically researching how to handle more practical oral instructions. Additionally, while the current framework maintains the dynamic nature of the environment when constructing the visual scene graph (e.g., the position of small objects), the room layout and all static objects remain unchanged after the scene graph is generated. This significantly limits the adaptability to new environments. Future work could explore how to automatically construct visual scene graphs in unknown environments. Lastly, although the framework is designed for dynamic environments, it primarily focuses on static objects, overlooking ongoing dynamic events in the environment. Addressing this gap would be a significant step towards the practical deployment of embodied robots in real-world settings.

\vspace{25pt}
\section{Conclusion}
In this work, we first present a new embodied complex-question answering task. Then, we propose an embodied task planning framework based on visual environment interaction for new task. By creating a structured semantic space, we can control the interaction between visual perception information and language chain expressions. This space enables our framework to generate sequential plans. Additionally, we present the ECQA dataset, which mainly enriches the template-based questions and adds multi-step questions. Experimental results show that our framework performs excellently on the new dataset and demonstrates superior performance in real-world validation, particularly for questions related to people and small objects.

\ifCLASSOPTIONcaptionsoff
  \newpage
\fi

\bibliographystyle{IEEEtran}
\bibliography{paper}

\begin{IEEEbiography}[{\includegraphics[width=1in,height=1.25in,clip,keepaspectratio]{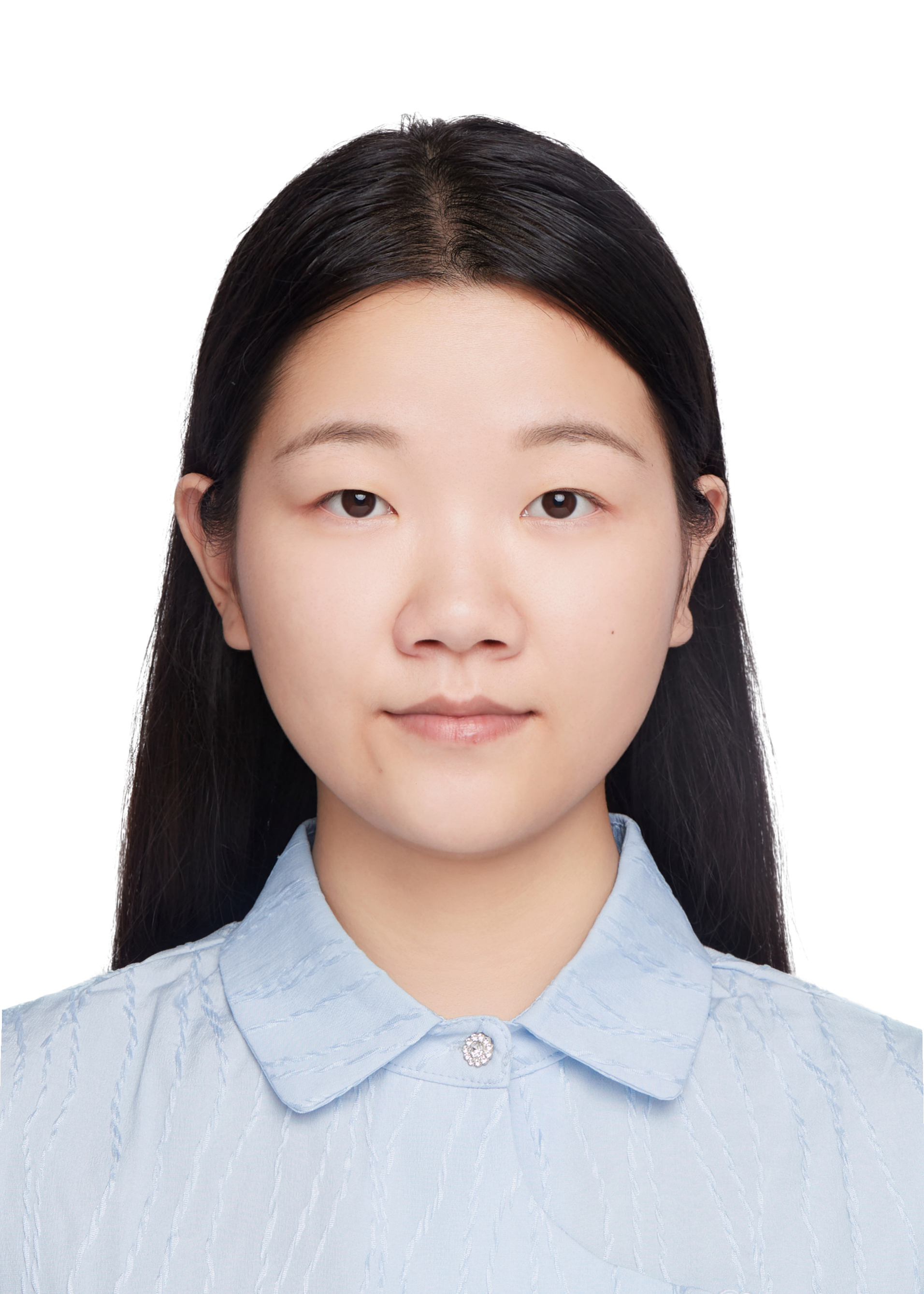}}]{Ning Lan}  received the M.S. degree from Shanghai University Of Engineering Science, Shanghai, China, in 2019. She is currently pursuing the Ph.D. degree at Xidian University, Xi'an, China. Her current research interests include Embodied intelligence, large language model, and computer vision.
\end{IEEEbiography}

\vspace{-12mm}
\begin{IEEEbiography}[{\includegraphics[width=1in,height=1.25in,clip,keepaspectratio]{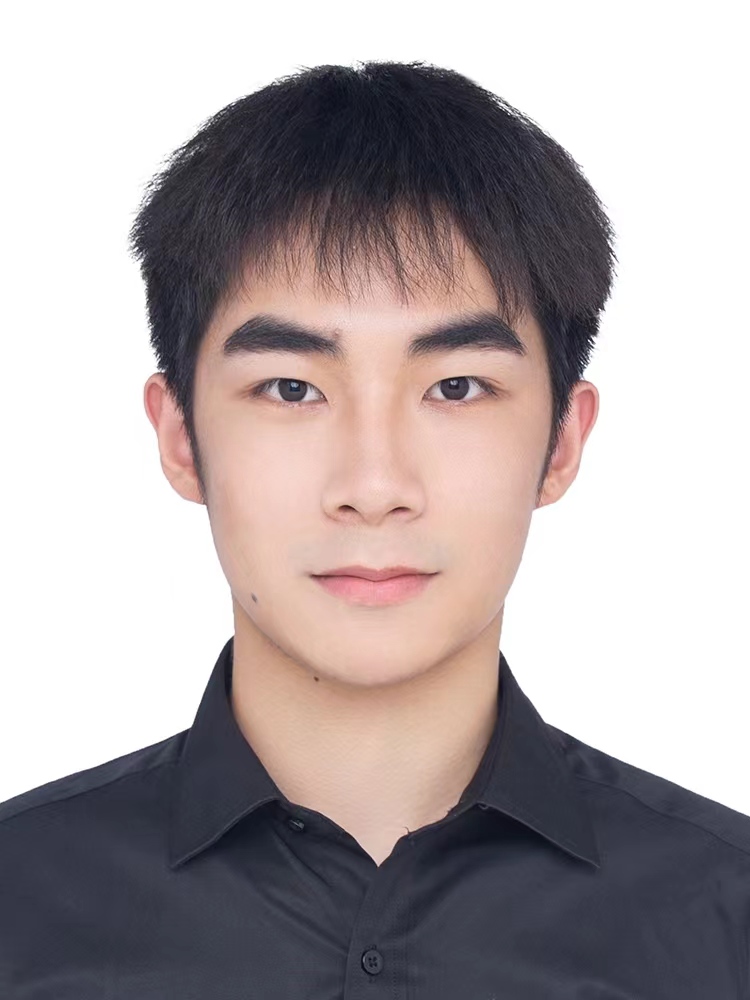}}]{Baoshan Ou}  received the B.S. degree in Intelligent Science and Technology from the University of Shanghai for Science and Technology, Shanghai, China, in 2023, and is currently pursuing the M.S. degree in Computer Technology at Xidian University, Xi’an, China. His research interests include embodied intelligence and large language model technology.
\end{IEEEbiography}
	
\vspace{-12mm}
\begin{IEEEbiography}[{\includegraphics[width=1in,height=1.25in,clip,keepaspectratio]{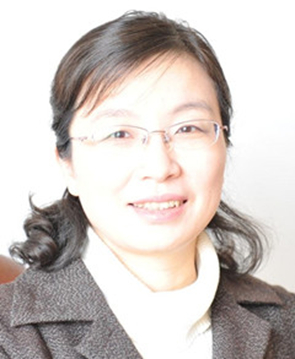}}]{Xuemei Xie} (Senior Member, IEEE) received the M.S. degree in electronic engineering from Xidian University, Xi’an, China, in 1994, and the Ph.D. degree in electrical and electronic engineering from The University of Hong Kong, Hong Kong, in 2004. She is currently a Professor with the School of Artificial Intelligence, Xidian University. She has authored more than 50 academic papers in international and national journals, and international conferences. Her research interests include artificial intelligence, compressive sensing, deep learning, image and video processing, and filter banks.
\end{IEEEbiography}
	
\vspace{-12mm}
\begin{IEEEbiography}[{\includegraphics[width=1in,height=1.25in,clip,keepaspectratio]{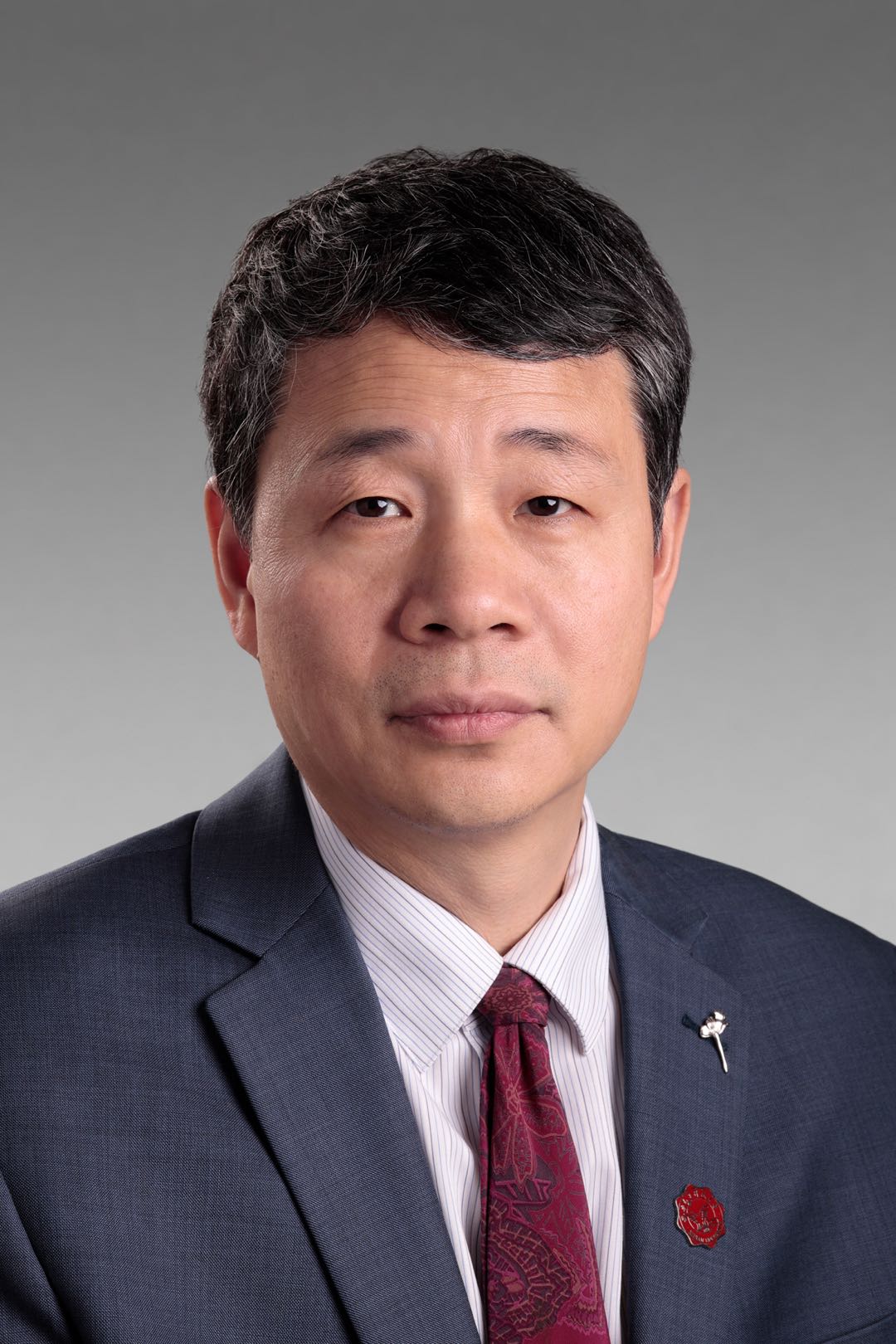}}]{Guangming Shi} (Fellow, IEEE) received the B.S. degree in automatic control, the M.S. degree in computer control, and the Ph.D. degree in electronic information technology from Xidian University in 1985, 1988, and 2002, respectively. From 1994 to 1996, he was a Research Assistant cooperated with the Department of Electronic Engineering, The University of Hong Kong. He joined the School of Electronic Engineering, Xidian University, in 1988. Since 2003, he has been a Professor with the School of Electronic Engineering, Xidian University. In 2004, he was the Head of the National Instruction Base of Electrician and Electronic (NIBEE). From June 2004 to December 2004, he studied with the Department of Electronic Engineering, University of Illinois at Urbana–Champaign (UIUC). He is currently the Deputy Director of the School of Electronic Engineering, Xidian University, and the Academic Leader in the subject of circuits and systems. He has authored or coauthored more than 100 research articles. His research interests include compressed sensing, the theory and design of multirate filter banks, image denoising, low-bit-rate image/video coding, and the implementation of algorithms for intelligent signal processing (using DSP and FPGA).
\end{IEEEbiography}

\end{document}